\documentclass[10pt,twocolumn,letterpaper]{article}

\usepackage[pagenumbers]{paper} 

\definecolor{wacvblue}{rgb}{0.21,0.49,0.74}
\usepackage[pagebackref,breaklinks,colorlinks,allcolors=wacvblue]{hyperref}

\title{An Evaluation Framework for Generating Multi-View Images of a Person in a Scene}

\author{
Mahir Majid \quad
Young Kyung Kim \quad
Guillermo Sapiro \\
Princeton University
}
\usepackage{booktabs} 
\usepackage{multirow}

\begin{document}
\twocolumn[{%
  \maketitle
  \vspace{0.1em} % small gap after authors (tune as needed)
  \begin{center}
    \includegraphics[width=0.95\textwidth]{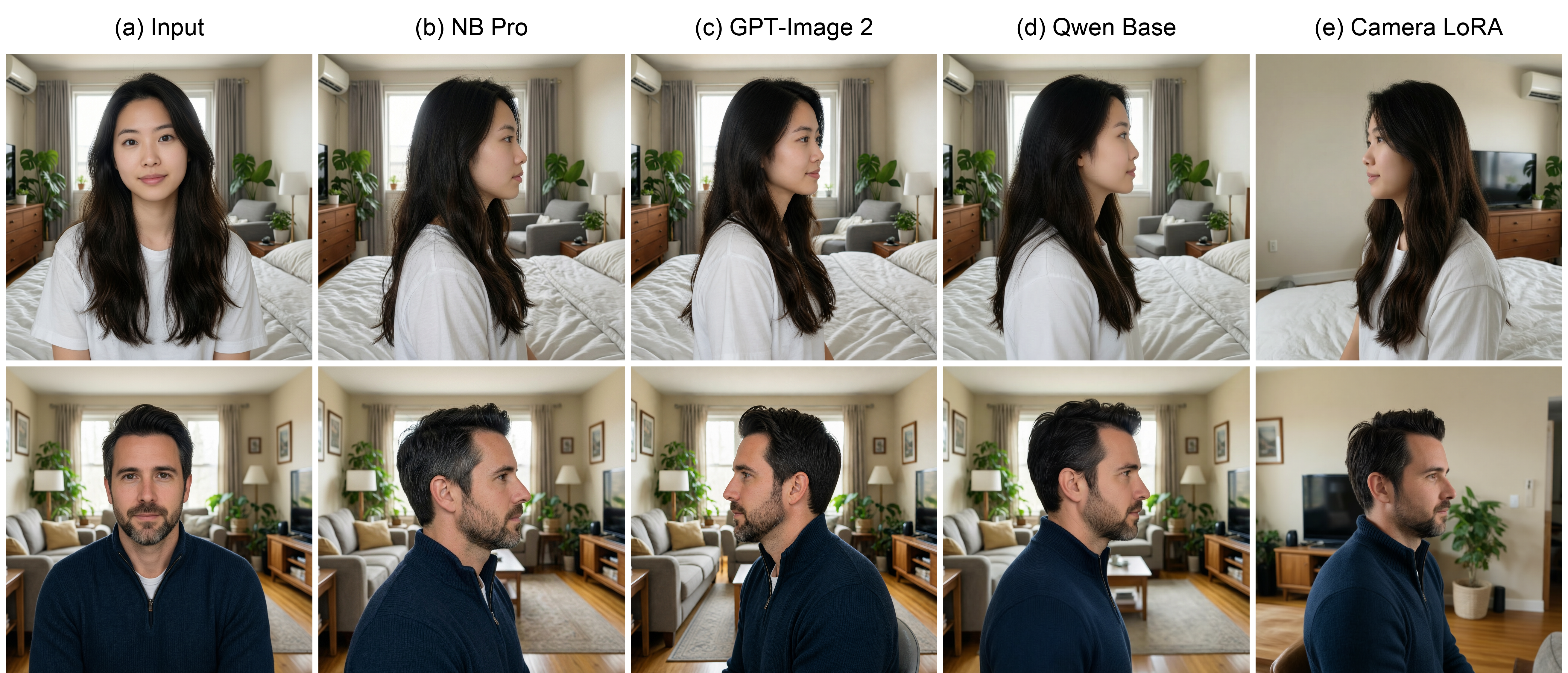}
    \captionof{figure}{\textbf{ Viewpoint editing across different models.} 
    (a) Input images of a woman in a bedroom (top row) and a man in a living room (bottom row). 
    Columns (b–d) show viewpoint-shifted outputs generated by baseline models with prompt variation P1 
    (defined in Section~\ref{sec:viewpoint_shift}), which tend to leave the background static. 
    Column (e) shows outputs from Qwen-Image-Edit-2511 with Camera LoRA. 
    The top row demonstrates a right view shift (bedroom) and the bottom row demonstrates a left view shift 
    (living room), illustrating how only Camera LoRA synthesizes correct perspective parallax and coherent scene 
    geometry changes alongside head pose alignment, while other models provide an inconsistent background.}
    \label{fig:viewpoint_shift_diagram}
  \end{center}
  \vspace{1em} % space before abstract
}]
\begin{abstract}
Recent generative image-editing Diffusion Transformers (DiTs) demonstrate impressive semantic editing capabilities but still struggle with spatially consistent camera angle changes. A primary bottleneck in training foundation models to execute free-form, promptable camera angle changes is the lack of specialized training data. While multi-view datasets exist for generic 3D environments and objects, there remains an absence of paired, multi-view datasets featuring human subjects at fixed locations in natural scenes, including frontal and side-profile views. Capturing such multi-camera data in unconstrained environments is logistically challenging and unscalable. In this paper, we first experiment with multiple state-of-the-art image editing models to create this data synthetically, but find that the outputs are frequently prone to hallucinations involving how much the subject's head turns relative to the background, often producing inconsistent environments. To address this issue, we propose the Head Scene Rotation Difference (HSRD) metric to quantitatively evaluate camera movements around a person. The proposed metric operates by decoupling camera movement from localized head pose manipulation. As demonstrated by the extensive experimentation, HSRD provides the pipeline necessary to evaluate 3D spatial parallax for a person in a scene, paving the way to reliably construct high-quality multi-view synthetic datasets. 

\end{abstract}
    
\section{Introduction}
\label{sec:intro}

The use of generative artificial intelligence for virtual production, filmmaking, and cinematic workflows increasingly relies on the ability to execute flexible camera orbits around human subjects. Recent advancements in image editing Diffusion Transformers (DiTs)~\cite{qwen2vl, flux2report2025, Feng2025, Huang2025} have demonstrated remarkable capabilities in instruction-driven semantic editing. However, despite their high-fidelity visual outputs, these foundation models fundamentally struggle with spatially rigorous novel-view synthesis. This limitation primarily stems from a lack of explicit 3D geometric priors in the underlying architecture, making geometrically consistent camera angle changes highly challenging without specialized, multi-view training data. 

Consequently, when these generative models are prompted to execute a large-angle camera rotation around a person in a natural scene, they frequently default to a structural path of least resistance. Instead of rendering a complex, geometrically accurate rotation of the background environment, the model often leaves the background rigid and unnaturally turns the subject's head to simulate the requested camera change. We illustrate this common failure, for multiple state of the art models, in Fig.~\ref{fig:viewpoint_shift_diagram}.

To reliably construct and curate training datasets containing multi-view images of humans in scenes, an evaluation pipeline is required to filter out these severe and very common hallucinations. While head pose analysis can identify changes in the subject's orientation, it is insufficient on its own. Instead, the analysis must jointly track changes in the subject's head pose and the surrounding scene geometry to determine whether the observed head pose change is consistent with a global camera transformation or results from an implausible localized deformation of the subject. Unfortunately, existing multi-view and 3D-consistency metrics~\cite{watson2023novel, yu2023photoconsistent, met3r2025, stern2026appreciateview} are ill-equipped for this specific task. Because contemporary 3D geometric metrics evaluate image consistency globally, they are not designed to mathematically decouple a legitimate global camera rotation from an inaccurate, localized head pose manipulation.

To address this critical gap, we propose the Head Scene Rotation Difference (HSRD), a novel evaluation metric designed to quantitatively evaluate camera movements around a human subject. By separating the background environment from the foreground subject, HSRD decouples the scene-based camera orbit from the localized head movement of the subject, providing a robust evaluation pipeline.
In summary, our main contributions are as follows:
\begin{itemize}
    \item We analyze the capability of existing state of the art image editing models for novel-view synthesis of scenes containing a human subject.
    \item We propose the HSRD evaluation metric and verify its reliability to measure consistency between head pose manipulation and the background scene change. 
    \item We demonstrate the efficacy of HSRD as a quality filtering mechanism on a synthetic dataset. 
\end{itemize}
\section{Related Work}
\label{sec:related}

\textbf{3D Human-Centric Datasets.} Training state-of-the-art foundation models to natively understand spatial camera orbits requires strictly paired multi-view training data~\cite{liu2023zero1to3, cameractrl2025}. Since base models are trained on unstructured, unpaired 2D datasets, their attention layers learn statistical approximations of varying angles rather than rigid mathematical camera transformations. Consequently, without explicit multi-view pairings to act as geometric constraints during training, these models fail to generate true 3D spatial parallax and instead hallucinate structural deformations~\cite{mvdream2024}. 

The current landscape of 3D-aware datasets lacks multi-view data of human subjects in scenes. On one end, scene-level datasets such as RealEstate10K~\cite{zhou2018stereo} and   CO3D~\cite{reizenstein2021common} provide rigorous ground-truth camera matrices and dense multi-view data, but they are strictly limited to isolated 3D environments or generic objects with no dynamic human subjects present. Conversely, multi-view human-centric datasets, such as MVHumanNet~\cite{zheng2024mvhumannet} and FaceScape~\cite{yang2020facescape}, capture humans with high geometric accuracy, but rely on synchronized multi-camera capture rigs. This restricts these datasets to controlled, uniform studio backgrounds and highlights a critical ``in-the-wild'' multi-view human dataset gap.

\noindent\textbf{Diffusion Transformers and Camera Angle Control.} Recent state-of-the-art image editing foundation models have increasingly transitioned toward DiT architectures~\cite{peebles2023dit, esser2024sd3, qwen2vl, flux2report2025}. While these architectures excel at complex semantic rendering and instruction-following, adapting them for spatially rigorous novel-view synthesis reveals a critical architectural bottleneck, namely their reliance on Rotary Position Embeddings (RoPE)~\cite{su2024roformer}. Since adjacent image patches in these models share continuous coordinate grids and retain significantly higher attention weights relative to one another, DiTs inherently suffer from a positional ``copy-bias.'' Recent mechanistic analyses of RoPE-based DiTs confirm this phenomenon, demonstrating that the high-frequency components of RoPE exert a disproportionately strong spatial bias that dominates attention computation, actively steering the network toward strict spatial alignment and ``reference copying'' rather than semantic transformation~\cite{untwistingrope2026, wei2025freeflux}. This spatial binding heavily restricts the massive global pixel shifts required to execute true camera angle changes, causing the base model to resist redrawing the background environment.

To circumvent this architectural restriction, a specialized camera angle change LoRA~\cite{hu2022lora} for Qwen-Image-Edit-2511~\cite{fal2026qwenangles} has been developed to force viewpoint changes synthetically. Since capturing perfectly paired and in-the-wild multi-view human data is logistically prohibitive, this type of adapter offers a highly promising avenue to generate the missing data. When successful, it can override the base model's 2D constraints to execute genuine logical camera angle changes around a person.

However, deploying this adapter at scale, as shown in the analysis in the following section, reveals that it is still highly prone to geometric hallucinations. Since a lightweight LoRA update must constantly fight against the base DiT's entrenched RoPE, which strictly binds output pixels to their original spatial coordinates, the network is still prone to turning only the head while keeping the background largely frozen or not executing a logically consistent scene change. 

Consequently, while this camera-control LoRA is capable of producing the desired novel views, its inconsistency means the raw output cannot be blindly trusted to curate geometrically sound 3D datasets. To safely construct synthetic training pairs, a filtering pipeline is required to mathematically separate the valid global camera orbits from the illicit localized head-turns. This systemic unreliability directly necessitates the robust, decoupled evaluation metric proposed in this work.

\noindent\textbf{Image Consistency Metrics.} Evaluating novel views has traditionally relied on 2D image-level metrics like LPIPS~\cite{zhang2018unreasonable} and SSIM~\cite{wang2004image}. However, since these strictly measure pixel-wise or perceptual variance, they cannot determine whether a massive pixel shift represents a mathematically valid 3D spatial rotation or a structural hallucination. Recently, a feed-forward 3D-consistency metric, MEt3R~\cite{met3r2025}, has emerged to evaluate dense geometric consistency without ground-truth poses. While effective for static environments, MEt3R processes the entire image holistically. By evaluating the full frame, its resulting geometric score fuses the foreground person and the background scene into a single, intertwined representation. Consequently, this metric is mathematically incapable of distinguishing a subject's localized pose change from the camera's global orbit. Because it fails to decouple the foreground human from the background, it cannot reliably determine whether a logically consistent novel view of the person has been generated. This motivates the development of the HSRD metric here introduced.
\section{Measuring Viewpoint Shift}
\label{sec:viewpoint_shift_analysis}

The purpose of this section is to measure the extent to which the background stays frozen when prompting different image editing base models for a camera angle change, a common error in SOTA models (Fig.~\ref{fig:viewpoint_shift_diagram}), and to test the extent to which utilizing a camera angle change LoRA addresses this failure point.

\subsection{Dataset Generation}
Initial attempts to parse existing segmentation repositories, such as Persons Supervisely \cite{supervisely_persons} and P3M-10k \cite{li2021privacy}, showed that standard in-the-wild datasets mostly feature people doing random actions rather than facing forward with a visible room behind them; this is much needed to re-orient people and a gap in current SOTA. To bypass this data constraint, we use the \texttt{Qwen-Image-2512} foundation model to generate our reference inputs. 
 
\begin{figure}[h]
  \centering
  \includegraphics[width=\linewidth]{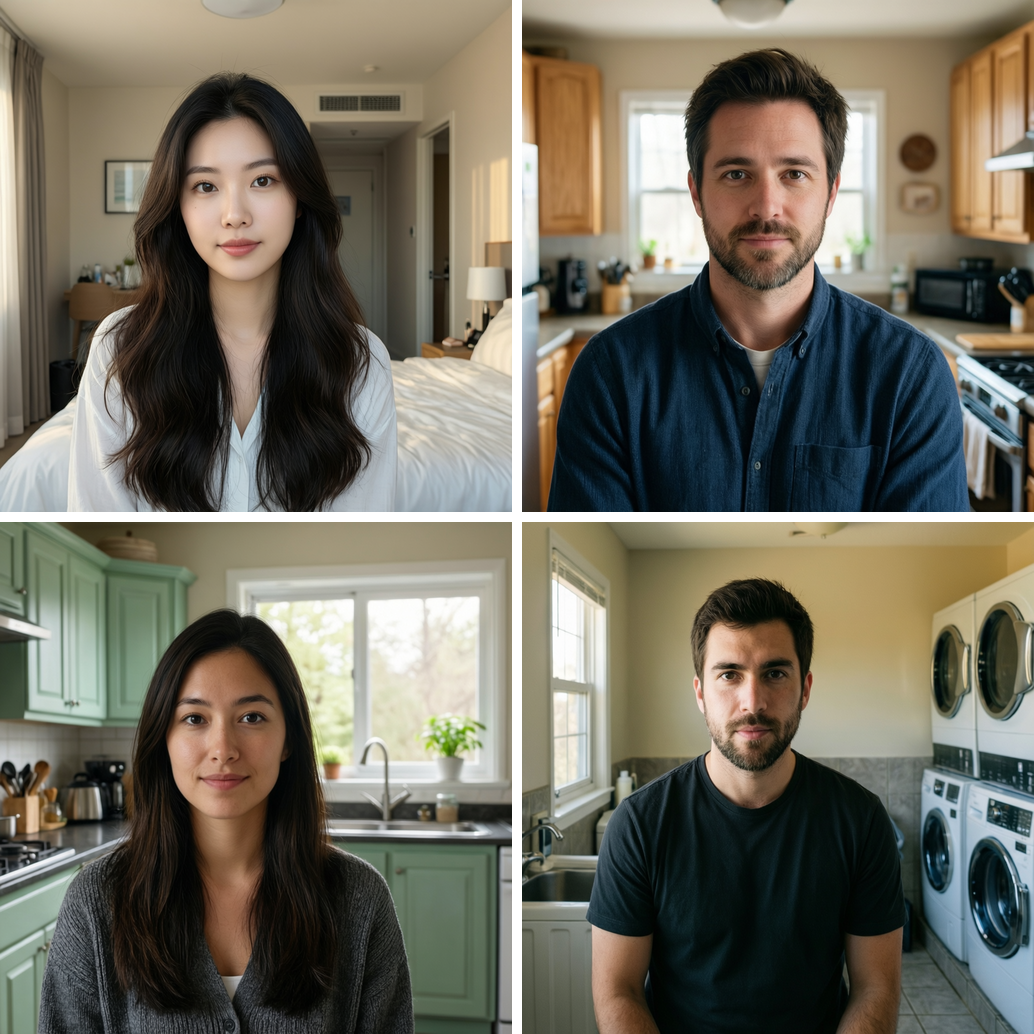}
  \caption{\textbf{Sample of frontal human scene dataset.} 
  A representative sample of the synthetic 40 image dataset generated from Qwen-Image-2512 including the following examples: (top left) Woman in Bedroom, (top right) Man in Kitchen, (bottom left) Woman in Kitchen, and (bottom right) Man in Laundry Room.
 }
  \label{fig:input_dataset_samples}
\end{figure}

To evaluate how the model handles complex contextual backgrounds while ensuring only a single person is present, we focus our dataset strictly on indoor residential scenes. Our proposed metric is, of course, more general than this. We define eight scene-gender categories combining male and female subjects across four core indoor settings: a bedroom, a living room, a kitchen, and a laundry room. For each category, we use the prompt structure:\footnote{The work here reported is independent of the critical problem regarding biases image generation models might have in creating and analyzing gender-related prompts.}\textit{``A [man/woman] in a [setting] looking forward, full room visible behind [him/her].''} By using 5 distinct generation seeds, we create a benchmark dataset of 40 frontal reference images. A sample of this dataset is shown in  Figure~\ref{fig:input_dataset_samples}. 

\subsection{Viewpoint Shift Generation}
\label{sec:viewpoint_shift}

\textbf{Base Model Edits.} We evaluate the following image editing base models: Qwen-Image-Edit-2511, Nano Banana Pro (Gemini Pro 3 Image), Nano Banana 2 (Gemini 3.1 Flash Image), and GPT Image 2. Additionally, we use 3 distinct textual prompt pairs describing directional viewpoint shifts to ensure our evaluation is robust to variations in natural language conditioning. Editing 40 images for both directional viewpoints across 3 prompt pairs resulted in 240 images per base model and 960 base images in total. The three prompt pairs used are:
\begin{enumerate}
    \item ``Prompt 1 (P1): A photorealistic (right/left) side profile view of the person. The camera perspective shifts to the (right/left) side of the person.''
    \item ``Prompt 2 (P2): A (right/left) side view of the person as if the camera has orbited around to their (right/left) side.''
    \item ``Prompt 3 (P3): Move the camera angle to the (right/left) side of the person.''
\end{enumerate}

\noindent\textbf{Camera Angle LoRA Edits.} We apply the specialized Multiple-Angles Camera LoRA \cite{fal2026qwenangles} at full strength ($1.0$) for the Qwen-Image-Edit-2511 model using the standardized trigger template: \texttt{<sks> [left/right] side view eye-level shot medium shot}. Each of the 40 reference images is edited for both left and right directions, yielding 80 generated camera LoRA outputs.

In total, the procedure involved 1040 generated image pairs from 40 starting Qwen-Image-2512 images. (All data in this project will be released upon acceptance.)

\subsection{Quantitative Results}

To evaluate how effectively a viewpoint change was executed, we measure both isolated pixel-level changes within the background using PSNR and MSE, and global perceptual shifts across the entire image using LPIPS. For isolating the background pixels from the input frontal image and the corresponding output image, we utilize the Segment Anything Model 3 (SAM 3)~\cite{carion2026sam3} to generate precise foreground segmentation masks and take the union for each paired evaluation. By applying the resulting background-exclusive segmentation mask, we compute masked PSNR and masked MSE.

If a model successfully rotates the camera, the background pixels should shift significantly, yielding a lower PSNR, a higher MSE, and a higher LPIPS score. Conversely, when a model fails and executes a localized head-turn—twisting the subject’s neck while locking down the surrounding background pixels—the metrics follow an inverse trend.

\begin{table}[t]
\centering
\caption{\textbf{Quantitative evaluation of background shift for detecting camera angle change in generated images.} Performance is aggregated across various models and prompt variations, as well as the Qwen camera angle change LoRA. Metrics are computed over a fixed evaluation subset of 80 generated images per group. Metric arrows indicate direction of lower background similarity.}
\label{tab:overall_results}
\begin{tabular}{@{}lccc@{}}
\toprule
\textbf{Configuration} & \textbf{PSNR $\downarrow$} & \textbf{MSE $\uparrow$} & \textbf{LPIPS $\uparrow$} \\
\midrule
GPT Image 2 (P1) & 22.8908 & 0.013336 & 0.4538 \\
GPT Image 2 (P2) & 22.2274 & 0.016670 & 0.4416 \\
GPT Image 2 (P3) & 13.6919 & 0.055676 & 0.5761 \\
\midrule
Nano Banana 2 (P1) & 27.4104 & 0.011071 & 0.4162 \\
Nano Banana 2 (P2) & 24.0676 & 0.024641 & 0.4469 \\
Nano Banana 2 (P3) & 22.4992 & 0.026416 & 0.4444 \\
\midrule
Nano Banana Pro (P1) & 28.7086 & 0.006930 & 0.3936 \\
Nano Banana Pro (P2) & 28.5215 & 0.008472 & 0.3840 \\
Nano Banana Pro (P3) & 25.8873 & 0.018506 & 0.4139 \\
\midrule
Qwen-2511 (P1) & 28.7264 & 0.003710 & 0.4080 \\
Qwen-2511 (P2) & 28.8497 & 0.002592 & 0.4025 \\
Qwen-2511 (P3) & 27.7679 & 0.005699 & 0.4122 \\
\midrule
\textbf{Qwen Camera LoRA} & \textbf{11.0014} & \textbf{0.095809} & \textbf{0.6447} \\
\bottomrule
\end{tabular}
\end{table}

As demonstrated in Table~\ref{tab:overall_results}, the Qwen-Image-Edit-2511 model with the camera LoRA demonstrates a more modified background scene on average than every other configuration including all of the prompt variations of GPT Image 2, Nano Banana 2, Nano Banana Pro, and Qwen-Image-Edit-2511. For instance, the average MSE for the camera LoRA is over four times higher than the average MSE of all the Nano Banana 2, Nano Banana Pro, and standard Qwen-Image-Edit-2511 configurations. This empirical evidence indicates the presence of Rotary Position Embedding (RoPE) positional bias, where the model anchors output pixels to their original spatial coordinates, effectively freezing the background and defaulting to a localized head-turn.

Furthermore, the impact of this bias can be qualitatively observed in \cref{fig:viewpoint_shift_diagram}. For two sample input images, it can be seen that the other base models keep the background from the input image largely the same while the camera LoRA is able to alter the background details sufficiently to present an appropriate camera angle change. It is also worth noting that the base models tend to naively use the lateral direction in the prompt as the direction in which the head should be facing for the output image. On the other hand, the camera LoRA can consistently use the lateral direction as the direction in which to move the camera, resulting in an output head pose that is opposite to the direction specified in the prompt, which is logically correct.  

The overall performance of the camera LoRA indicates that it is significantly more reliable than the existing base models for synthesizing multi-view data of humans in scenes. This points to the importance of curating multi-view data for enabling models to perform camera angle changes. However, to quantify how often the camera angle change LoRA outputs succumb to not changing the background, we define a failure mode, ``approaching base performance,'' as any LoRA output score falling within one standard deviation ($1\sigma$) of any base model's metric distribution. Our statistical analysis reveals that 12.25\% of all camera LoRA generations fail to overcome this innate positional anchoring, triggering the exact background-freezing structural failure observed in the base models. 

While the metrics used in this procedure can analyze whether a model has attempted a viewpoint transformation, they remain fundamentally limited as global proxies. Since they measure overall scene and background variance without explicit geometric constraints, they cannot distinguish between an arbitrary pixel distortion and a mathematically valid 3D spatial rotation of the background. Consequently, verifying whether a genuine, geometrically consistent camera orbit has occurred requires a specialized spatial protocol, which will be addressed by the proposed HSRD metric.
\section{Methodology: HSRD Evaluation Protocol}
\label{sec:methodology}

To rigorously evaluate whether a generative image editing DiT has executed a valid novel view synthesis or simply defaulted to a localized head rotation failure, an evaluation metric must separate foreground manipulation from background geometry. Existing 3D consistency metrics are computationally blind to this distinction because they compute a single fused geometric score across the entire unmasked frame. 

To address this critical gap, we introduce the HSRD protocol. HSRD acts as an evaluation pipeline that quantifies generative camera angle changes by mathematically decoupling the camera orbital movement from the localized head pose manipulation of the subject. The protocol, which is independent of the model used to generate the scene, is executed in three distinct stages: Absolute Head Pose Extraction, Background Isolated Camera Pose extraction, and the final Decoupled Rotation Difference calculation.

\subsection{Absolute Head Pose Extraction}
\label{subsec:head_pose}

To quantify head orientation, we deploy 6DRepNet~\cite{hempel20226drepnet}, an architecture optimized for 3D head pose estimation under extreme rotation angles. Facial bounding boxes are identified using MTCNN face detection~\cite{zhang2016joint} and cropped prior to pose prediction. 6DRepNet estimates the absolute 3D Euler angles—specifically yaw, pitch, and roll—for both the reference input image and the generated output image. 

By comparing the initial head pose to the output head pose, we calculate the absolute Head Pose Yaw Difference ($\Delta \text{Yaw}_{\text{head}}$). This value serves as a quantitative proxy for the localized rotation of the subject neck.

\subsection{Background-Isolated Camera Pose Estimation}
\label{subsec:background_isolated}

When a human subject occupies a portion of an image frame, their dynamic geometry and visible facial features can bias feed-forward 3D reconstruction models. To ensure that calculated camera motions reflect true environmental orbital movement rather than physical subject details, we ensure the human pixels are not presented to VGGT~\cite{vggt2025}, which is the 3D foundation model used for camera pose estimation.

First, we utilize SAM 3~\cite{carion2026sam3} with the text prompt \texttt{"person"} to generate binary foreground segmentation masks $M_1, M_2 \in \{0, 1\}^{H \times W}$ for the original reference input image $I_1$ and the generated viewpoint-shifted output image $I_2$, respectively. For each mask $M_k(u,v)$ --- where $k \in \{1, 2\}$) --- a value of $1$ denotes a foreground human pixel, and $0$ denotes a background pixel. Before passing these images to VGGT, we suppress the human foreground by setting its pixels to black in both views. This enables the construction of the masked image pair $(\tilde{I}_1, \tilde{I}_2)$, where $(I_1, I_2)$ represent the original image pair, via
\begin{equation}
\tilde{I}_k(u,v) = I_k(u,v) \cdot \big(1 - M_k(u,v)\big).
\label{eq:masking}
\end{equation}
This pixel-space suppression removes foreground features that could bias the pose estimation, while maintaining the contiguous spatial grid of the image patch tokens during VGGT's forward pass. Since VGGT is a global multi-view reconstruction transformer designed to estimate camera geometry within a shared spatial coordinate system, it calculates absolute camera poses relative to a common origin rather than computing relative pairwise transformations directly. Consequently, for a pair of input images, the network models a function $f_{\text{VGGT}}$ that maps the masked pair to their respective absolute camera rotation matrices $R_1, R_2 \in SO(3)$ in this shared coordinate system:
\begin{equation}
(R_1, R_2) = f_{\text{VGGT}}(\tilde{I}_1, \tilde{I}_2).
\label{eq:vggt_forward}
\end{equation}
To find the relative camera rotation matrix $R_{12} \in SO(3)$ from the
reference view to the edited view, we perform a coordinate change-of-basis.
Here, $R_1$ and $R_2$ transform 3D directions from the shared world
coordinate frame into the reference and edited camera frames, respectively.
Thus, to transform a direction from the reference camera frame to the edited
camera frame, we first apply $R_1^T$ to recover the world-frame direction,
followed by $R_2$ to express it in the edited camera frame:

\begin{equation}
R_{12} = R_2 R_1^T.
\label{eq:relative_rotation}
\end{equation}
To isolate the horizontal camera panning trajectory, we decompose $R_{12}$
into Euler angles representing yaw ($\theta_{\text{yaw}}$), pitch
($\theta_{\text{pitch}}$), and roll ($\theta_{\text{roll}}$). The horizontal
camera yaw $\theta_{\text{yaw}}$ is extracted from the relative rotation
matrix $R_{12}$ as
\begin{equation}
\Delta \text{Yaw}_{\text{scene}} = \theta_{\text{yaw}}
= \operatorname{atan2}\big(-R_{12}^{(2,0)}, R_{12}^{(2,2)}\big).
\label{eq:yaw_extraction}
\end{equation}
where $R_{12}^{(i,j)}$ denotes the element at row $i$ and column $j$ using
0-indexed notation. This formulation extracts the horizontal angular
displacement from the orientation of the camera's forward axis, providing a
yaw measurement corresponding to the camera's horizontal panning trajectory.

While VGGT is highly robust on standard uncorrupted image pairs, its geometric reliability when processing frames with blacked-out human occlusions has not been explicitly established in prior literature. In Section~\ref{sec:experiments}, we empirically validate this protocol on RealEstate10K, proving that VGGT reliably recovers ground-truth camera pose with negligible error under background-isolated alignment from a diverse set of human-shaped masks. Furthermore, VGGT consistently outperforms 3D foundational baselines like MASt3R and DUSt3R, justifying our architectural choice for this evaluation pipeline. 

\subsection{Calculating Decoupled Rotation Difference}
\label{subsec:calculating_hsrd}

The final step of the HSRD protocol establishes an objective decision rule to detect generative hallucinations by comparing the local head movement against the global camera movement. This is quantified using the Head Scene Rotation Difference (HSRD) metric.

In standard horizontal camera orbits, the camera's pitch and roll remain relatively stable. Consequently, the HSRD calculation proceeds under the geometric assumption that changes in the roll and pitch for both the head and the scene are negligible ($\Delta \text{Pitch} \approx 0$ and $\Delta \text{Roll} \approx 0$). 

Focusing strictly on the horizontal azimuth, the HSRD is defined mathematically as

\begin{equation}
\displaystyle
\text{HSRD} = \left| \Delta \text{Yaw}_{\text{head}} - \Delta \text{Yaw}_{\text{scene}} \right|.
\end{equation}

\begin{itemize}
    \item \textbf{True Viewpoint Synthesis.} If a model successfully moves the camera around a static subject, the scene yaw ($\Delta \text{Yaw}_{\text{scene}}$) will reflect the large spatial shift. Because the subject remains looking straight ahead relative to their own body, the change in head yaw relative to the new camera angle tracks identically to the camera shift. This results in a relatively lower HSRD, verifying a structurally sound camera orbit.
    
    \item \textbf{Intermediate Viewpoint Synthesis.} If the model attempts a partial camera rotation but simultaneously twists the subject's head to compensate for the remaining prompted angle, the scene and head yaws will decouple. This geometric mismatch results in a relatively higher HSRD, exposing an incomplete spatial orbit.
    
    \item \textbf{Localized Head Rotation Failure.} If the text prompt requests a camera angle change but the model succumbs to positional bias, the background remains rigidly locked in place ($\Delta \text{Yaw}_{\text{scene}} \approx 0$). However, to satisfy the semantic request of the prompt for a profile view, the model will likely twist the head of the subject resulting in a high $\Delta \text{Yaw}_{\text{head}}$ and a massive spike in the HSRD score.
\end{itemize}
\section{Experimental Results}
\label{sec:experiments}

\subsection{Validating HSRD}
\label{subsec:validating_hsrd}

\textbf{Scene Dataset Selection.} A primary challenge in benchmarking human-centric novel-view synthesis is the absence of in-the-wild datasets that provide paired, multi-view images of dynamic humans alongside ground-truth camera pose matrices. Since this ideal dataset does not exist, we cannot directly verify the baseline reliability of 3D feed-forward models on actual humans in natural scenes. Instead, we must validate our protocol by proving that artificially masking out human shaped pixels from the background environment does not break the underlying 3D camera pose estimation. 

To validate our approach, we conduct experiments using the RealEstate10K~\cite{zhou2018stereo} dataset, a massive collection comprising nearly ten million frames derived from eighty thousand diverse video walkthroughs. The spatial data in this benchmark is highly precise, with camera extrinsics meticulously calculated through Simultaneous Localization and Mapping (SLAM) and optimized via bundle adjustment~\cite{mast3r2024}. Since these environments consist of predominantly static architectures, they offer an uncorrupted control setting. This enables us to show that removing foreground subjects via artificial masking does not degrade the core geometric tracking capabilities of the underlying 3D feed-forward model.

\begin{table*}[t]
\centering
\caption{\textbf{Quantitative impact of foreground segmentation on camera pose estimation.} Evaluation of background tracking stability on the RealEstate10K dataset. Errors are reported in degrees. Metric arrows denote direction of better performance.}
\label{tab:masking_results}
\begin{tabular}{llcccc}
\toprule
\textbf{Model} & \textbf{Mode} & \textbf{AUC@30 ($\uparrow$)} & \textbf{Yaw Error ($\downarrow$)} & \textbf{Pitch Error ($\downarrow$)} & \textbf{Roll Error ($\downarrow$)} \\
\midrule
VGGT   & \texttt{Original (Unmasked)}    & 88.33 & 1.21$^\circ$ & 0.52$^\circ$ & 0.98$^\circ$ \\
VGGT   & \texttt{Foreground Masking}     & 86.93 & 1.54$^\circ$ & 0.66$^\circ$ & 1.11$^\circ$ \\
\midrule
MASt3R & \texttt{Original (Unmasked)}    & 77.63 & 7.05$^\circ$ & 2.31$^\circ$ & 2.15$^\circ$ \\
MASt3R & \texttt{Foreground Masking}     & 76.15 & 8.29$^\circ$ & 2.37$^\circ$ & 2.22$^\circ$ \\
\midrule
DUSt3R & \texttt{Original (Unmasked)}    & 71.19 & 9.76$^\circ$ & 2.98$^\circ$ & 2.14$^\circ$ \\
DUSt3R & \texttt{Foreground Masking}     & 69.15 & 10.63$^\circ$ & 3.26$^\circ$ & 2.16$^\circ$ \\
\bottomrule
\end{tabular}
\end{table*}

\noindent\textbf{Human Mask Validation Set.} To simulate realistic human occlusions within the RealEstate10K scenes, we leverage the validation split of P3M-10k~\cite{li2021privacy}, which contains $1,000$ high-resolution human portrait matte masks. We select P3M-10k because its fine-grained alpha mattes preserve intricate human boundary details including hair strands and clothing contours, providing a diverse set of high-fidelity human foreground masks that prevent boundary artifacts during segmentation. 

\noindent\textbf{Evaluation Metrics.} Following established evaluation procedures on RealEstate10K~\cite{vggt2025, wang2023posediffusion}, we randomly sample 10 frames per video sequence and compute the estimated camera shifts across all possible paired combinations within that sequence. Overall geometric stability is quantified via the Area Under the Curve at a $30^\circ$ threshold (AUC@30), an aggregate metric incorporating both Relative Rotation Accuracy (RRA) and Relative Translation Accuracy (RTA) to measure angular deviation and spatial displacement.

In addition to this standard composite score, we explicitly isolate and compare horizontal yaw error along with pitch and roll error across all evaluation conditions. Because the fundamental decision rule of our proposed HSRD metric relies strictly on decoupling head yaw from camera yaw, verifying that our masking technique preserves background yaw stability is paramount. 

To evaluate the structural impact of human-shaped foreground masking on 3D geometric tracking, we benchmark 2 distinct modes on the 3D feed-forward models VGGT, MASt3R, and DUSt3R:

\begin{enumerate}
    \item \textbf{Original.} The original RGB image pair is evaluated without any image modification or masking.
    \item \textbf{Foreground Masking.} For each image pair, two human-shaped masks are randomly drawn from P3M-10k to zero out pixels in both images. The model predicts the camera pose over the union mask of the image pair.
\end{enumerate}

\noindent\textbf{Results.}
As detailed in Table~\ref{tab:masking_results}, VGGT consistently outperforms both MASt3R and DUSt3R across all tracking metrics, establishing it as the most robust 3D foundation model for our evaluation pipeline. Additionally, the geometric performance remains stable between the original unmasked baseline and the foreground masking mode, confirming that artificial human-shaped pixel zeroing is not fundamentally detrimental to the scene reconstruction. 

For instance, the performance of VGGT under foreground masking still outperforms the models MASt3R and DUSt3R on the unmasked baseline, indicating that the foreground masking procedure under VGGT is still reliable for pose estimation. 
Thus, the proposed methodology of calculating HSRD by blacking out foreground pixels, to ensure the dynamic features of the human subject have no impact on VGGT's pose estimation, is reasonable and the model is still able to recover an accurate background change. 

\subsection{HSRD Filtering for Multi-View Human Scene Data Synthesis} \label{subsec:hsrd_filtering}

\noindent \textbf{Dataset Preparation.} As discussed in Section~\ref{sec:viewpoint_shift_analysis}, the usage of the camera LoRA with Qwen-Image-Edit-2511 was significantly more reliable than the other base models in generating multi-view images of human subjects in scenes since the base models were prone to naively freezing the background. Thus, to showcase the usefulness of HSRD as a quality filter, we create a synthetic dataset with the camera LoRA following a similar procedure to that used for the camera LoRA outputs in Section~\ref{sec:viewpoint_shift_analysis}, using five distinct seeds to produce a synthetic dataset of 400 camera LoRA outputs.

\noindent \textbf{Results.} Table~\ref{tab:quartile_variance} illustrates the quartile distribution of the HSRD metric across the overall dataset, as well as breakdown by lateral view directions. Across the entire dataset, HSRD exhibits an exceptionally high variance and a wide distributional range. This dispersion indicates that a substantial portion suffers from severe spatial entanglement.

\begin{table}[h]
\centering
\caption{\textbf{HSRD distribution and quartile analysis by view direction.} Breakdown of HSRD metrics across the overall dataset and directional subsets, highlighting the skew toward higher rotational discrepancy in right-ward prompts.}
\label{tab:quartile_variance}
\resizebox{\columnwidth}{!}{%
\begin{tabular}{lcccc}
\toprule
\textbf{Subset} & \textbf{Mean / Std} & \textbf{$Q_1$} & \textbf{Median} & \textbf{$Q_3$} \\
\midrule
\textbf{All Views}   & $47.05^\circ \pm 44.83^\circ$ & $8.70^\circ$ & $30.52^\circ$ & $73.35^\circ$ \\
\textbf{Left Views}  & $34.72^\circ \pm 35.35^\circ$ & $6.35^\circ$ & $16.61^\circ$ & $61.38^\circ$ \\
\textbf{Right Views} & $59.37^\circ \pm 49.66^\circ$ & $14.20^\circ$ & $63.64^\circ$ & $84.56^\circ$ \\
\bottomrule
\end{tabular}%
}
\end{table}

\begin{figure*}[t]
    \centering
    \includegraphics[width=0.95\textwidth]{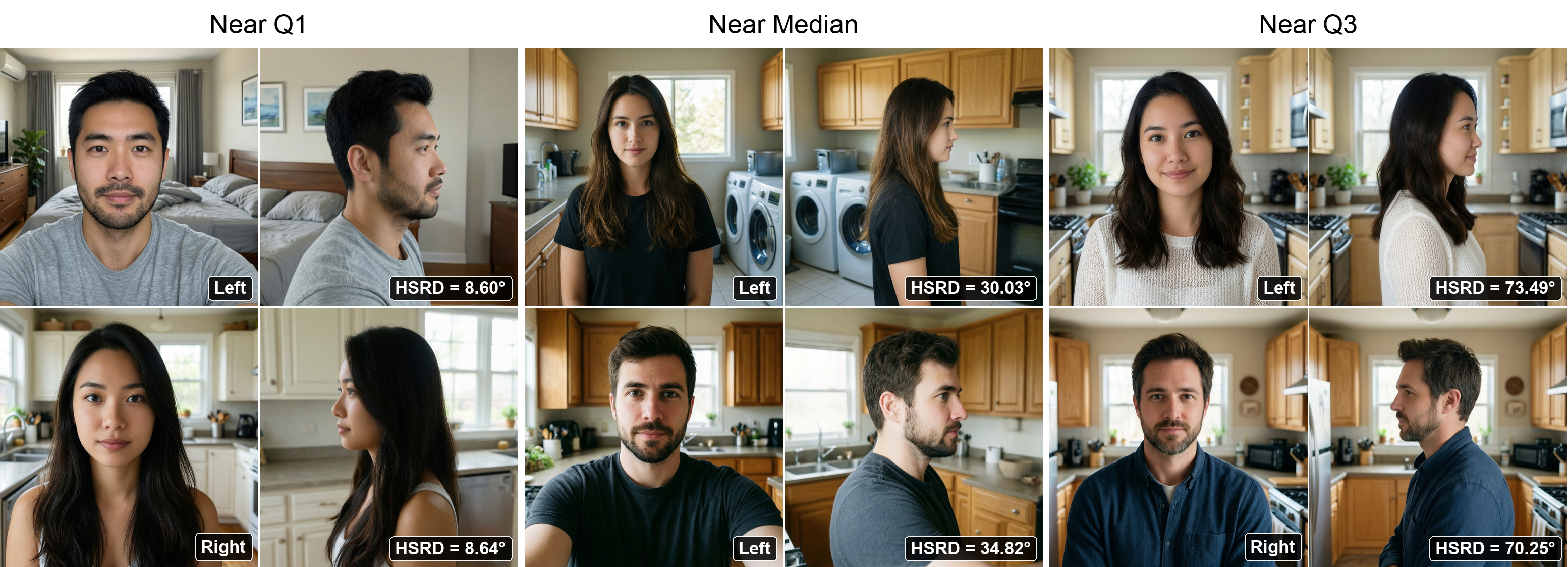}
    \caption{\textbf{HSRD analysis of camera LoRA output images.} Input images on the left for each pair are labeled with the requested camera change direction, such that the head in the output image should face opposite to the labeled direction. LoRA output images are labeled with their corresponding HSRD scores. Images with lower HSRD scores show logically consistent camera angle changes, while images with higher HSRD scores near Q3 tend to exhibit a frozen background with only the head turning to simulate the requested camera change.}
    \label{fig:hsrd_visual_thresholds}
\end{figure*}

Beyond these quantitative distributions, qualitative visual analysis, as shown in \cref{fig:hsrd_visual_thresholds}, reveals a clear correlation between HSRD scores and geometric fidelity of the camera angle change. Generations with notably low HSRD scores, clustering near the first quartile ($Q_1$), exhibit highly logical and mathematically sound camera orbits. 

For instance, under a leftward requested view shift for the top example of the man in the bedroom under the `Near Q1' section, the bed appears at the correct relative angle and objects including the pillows as well as the painting are logically positioned in the output image. Similarly, the bottom example for the `Near Q1' section of the woman in the kitchen demonstrates a successful rightward requested view shift. The window and counter, originally positioned on the left side of the frontal input, are correctly repositioned in the LoRA output, while previously unobserved regions, such as the newly synthesized cabinets to the left of the window in the output image, are also placed logically. These examples illustrate that low HSRD values correspond to outputs in which the background and human subject undergo a coherent spatial transformation.

Conversely, image pairs in \cref{fig:hsrd_visual_thresholds} scoring near the median HSRD display inconsistent performance. The top example in the `Near Median' section of the woman in the laundry room showcases a passable example of a leftward requested camera shift. However, for the bottom example of a leftward requested view shift of the man in the kitchen, there is an incomplete spatial orbit. Unlike in the input image where the window is directly facing the human subject's back, this is no longer so clear in the output image where the window seems to be facing the person's back at an angle, suggesting a stunted camera orbit. Guided by the clear geometric divide between structurally logical low-error orbits and inconsistent outputs, we establish $\text{HSRD} \le 20^\circ$ as a reasonable threshold for isolating valid novel-view synthesis.

Furthermore, the examples in \cref{fig:hsrd_visual_thresholds} under the `Near Q3' section showcase the reliability of HSRD in detecting outputs with background freeze where the model will hardly change the background, but severely rotate the person's head to simulate the requested camera angle change. As established earlier, such outputs are common especially with the base models and the fact that these images cause a high spike in the HSRD score is useful in filtering out these hallucinations. The sharp increase occurs because the HSRD procedure detects approximately zero rotation change in the background, but a significant rotation change of the head between the input image where it is facing forward and the output image where it is primarily facing one side.

However, since HSRD jointly tracks the consistency between the rotation of the human's head and the scene, a ``False Zero'' can occur if the model entirely ignores the prompt and does absolutely nothing. For instance, a $0^\circ$ head turn and $0^\circ$ scene turn yields an HSRD of 0. Another way a ``False Zero" can happen is if the output contains a logically consistent camera angle change, but in the opposite direction that was requested. Therefore, a valid multi-view synthesis must jointly exhibit low HSRD \textit{and} an accurate head pose change. We establish a minimum head pose shift threshold of $\ge 30^\circ$ in the expected direction to guarantee that a meaningful geometric transition from a frontal view to a side profile actually occurred. Across the entire dataset, 74.25\% of all generated images successfully rotated the head by at least $30^\circ$ in the requested direction. 

Out of the 400 total generations, 176 images fell below our strict $20^\circ$ HSRD threshold. Of these generations, 169 images (96\%) simultaneously passed the $\ge 30^\circ$ head pose change threshold where the head turned in the correct direction. The remaining 7 outliers represent the aforementioned ``False Zero'' generations. Ultimately, HSRD filtered the original dataset of 400 images to yield 169 geometrically valid multi-view image pairs.
\section{Conclusion}

The proposed here HSRD metric provides a useful method for evaluating generative novel-view synthesis for scenes involving a human subject. Since it effectively isolates valid camera orbits from localized head-turn failures, HSRD can function as a reliable filtering pipeline for creating synthetic multi-view data of a person in a scene. By generating large volumes of synthetic images and filtering out those with high HSRD scores, researchers can automatically curate the massive, geometrically clean multi-view datasets of humans in scenes that the AI filmmaking industry requires.

Lastly, the HSRD methodology is designed to evaluate a single primary human subject in a scene. Future work might extend this evaluation to scenes containing multiple primary human subjects and people present in the background, which is necessary to fully validate these models for real-world cinematic use.
\section*{Acknowledgments}

The authors are partially supported by ONR, NSF, and Apple.
{
    \small
    \bibliographystyle{ieeenat_fullname}
    \bibliography{main}
}

\end{document}